\documentclass[11pt,a4paper]{article}

\usepackage[hyperref]{acl2020}
\usepackage{times}
\usepackage{graphicx}
\usepackage{amsfonts}

\usepackage{xcolor}
\usepackage{amsmath}
\usepackage[linesnumbered,ruled,vlined]{algorithm2e}
\usepackage{latexsym}
\usepackage{threeparttable}

\usepackage{microtype}

\aclfinalcopy 

\SetCommentSty{mycommfont}

\SetKwInput{KwInput}{Input}                
\SetKwInput{KwOutput}{Output}

\title{DialBERT: A Hierarchical Pre-Trained Model for Conversation Disentanglement}

\author{ Tianda Li$^1$, Jia-Chen Gu$^2$, Xiaodan Zhu$^1$, Quan Liu$^{2,3}$,Zhen-Hua Ling$^2$, Zhiming Su$^3$, Si Wei$^3$\\
  $^1$ECE \& Ingenuity Labs, Queen's University, Kingston, Canada \\
  $^2$National Engineering Laboratory for Speech and Language Information Processing, \\
      University of Science and Technology of China, Hefei, China \\
  $^3$ State Key Laboratory of Cognitive Intelligence, iFLYTEK Research, Hefei, China \\
{\tt \{tianda.li,xiaodan.zhu\}@queensu.ca}, {\tt gujc@mail.ustc.edu.cn}, \\ {\tt \{quanliu,zhling\}@ustc.edu.cn},  {\tt \{zmsu,siwei\}@iflytek.com} 
}
\date{}
\begin{document}
\maketitle
\begin{abstract}
Disentanglement is a problem in which multiple conversations occur in the same channel simultaneously, and the listener should decide which utterance is part of the conversation he will respond to. 
We propose a new model, named Dialogue BERT (DialBERT), which integrates
local and global semantics in a single stream of messages to disentangle the conversations that mixed together. 
We employ BERT to capture the matching information in each utterance pair at the utterance-level, and use a BiLSTM to aggregate and incorporate the context-level information.
With only a 3\% increase in parameters, a 12\% improvement has been attained in comparison to BERT, based on the F1-Score. The model achieves a state-of-the-art result on the a new dataset proposed by IBM  \cite{kummerfeld-etal-2019-large} and surpasses previous work by a substantial margin.
\end{abstract}


\section{Introduction}

With the growth of the Internet, single stream conversation occurs everyday. But in many cases, all messages are entangled with each other, which makes it difficult for a new user to the chat in understanding the context of discussion in the chat room. Automatic disentanglement can help people leverage this type of problem. Research relevant to disentanglement has been conducted over a decade. 
 \citet{aoki2003mad,aoki2006s} attempted to disentangle conversation speech. Then, several studies were conducted in text-based conversational media. 
 The primary mainstream strategy of solving disentanglement is modeled as the Topic Detection and Tracking (TDT) task \cite{allan2002introduction}, which calculates sentence pair similarity iteratively and decides which conversation the message belongs to.
In addition to neural networks,  statistical \cite{du2017discovering} and linguistic features \cite{elsner-charniak-2008-talking,elsner-charniak-2010-disentangling,Elsner:2011:DCL:2002472.2002622,Mayfield:2012:HCS:2392800.2392810} are also used in existing research. Without considering relationships between words in a message, the quality of features will affect the final result dramatically.  Most previous works either used small dataset \citep{elsner-charniak-2008-talking} or used an unpublished dataset \citep{inproceedings,jiang2018learning}.

The publication of a new large-scale dataset \cite{kummerfeld-etal-2019-large} made it possible to train a more complex model. \citet{DBLP:conf/aaai/ZhuNWNX20} proposed masked hierarchical transformer based on BERT to learn a better conversation structure. But their approach need conversation structure in advance to formulate the Mask for their transformer model, which is not possible during the test process.

Unsupervised pre-training \cite{peters2018deep,radford2018improving,devlin2018bert,DBLP:journals/corr/abs-1906-08237,DBLP:journals/corr/abs-1907-11692,sanh2019distilbert,lan2019albert} has recently been shown to be very effective
in improving the performances of a wide range
of NLP tasks \cite{li2019several}.
Previous work has also explored the possibility of post-training BERT in the Ubuntu response selection task \cite{whang2019domain}.

In this paper, we propose our own pre-trained model, named Dialogue BERT (DialBERT). Conversation disentanglement leverages DialBERT in three steps: (1) Using BERT to capture the matching information in each sentence pair. (2) Using a context-level BiLSTM to incorporate the context information. (3) Measuring the semantic similarities between sentence pairs. Selecting the pair with the highest score and classifying this pair into the same conversation.
The usage of context-level BiLSTM enables BERT to capture semantics across a wide range of contexts. With only  a 3 \% increase in parameters, DialBERT shows an improvement on all evaluation metrics including an improvement of 12\% on the F1-score in comparison to BERT \cite{devlin2018bert}. Also, our model surpasses previous baseline models according to all the evaluation metrics by a wide margin, becoming a new state-of-the-art model. 

\section{Problem Statement}
In this section, we formally define the object of this work and notations used.

Every message  can be defined as a tuple $m$ = ($u$, $s$, $t$), where $u = \left \langle u_{1}, u_{2}, ..., u_{n} \right \rangle$ is the word sequence posted by speaker $s \in S$  at time $t$, where $S$ is the set of speakers. Each message belongs to a specific conversation $A$. Messages from different conversations could occur in the same channel concurrently.
Every message has a preceding message, which is an indication of the message being the response to its respective preceding message. 

Following the setting of previous works \cite{elsner-charniak-2008-talking,elsner-charniak-2010-disentangling,Elsner:2011:DCL:2002472.2002622,Mayfield:2012:HCS:2392800.2392810,jiang2018learning}. Our objective for model training is to calculate the similarity between two messages and to identify which message precedes the current message based on the calculated similarity. As \citet{jiang2018learning} pointed out, it is not required to calculate all message pair similarities to find out the preceding message. In our work, $T-1$ messages occurring before this target message are taken in to consideration.  Target message will be represented by $M = \left\{ m_{1}, m_{2},..., m_{n} \right\}(M \in \mathbb{R}^{n})$, and context messages will be represented by $C_{t}= \left\{ c^{t}_{1}, c^{t}_{2},..., c^{t}_{m}\right\}$ $(C \in \mathbb{R}^{T\times m})$, where $n$ and $m$ are the sequence length of messages and $t$ is the index of the preceding message. $Y \in \left\{ 0, 1\right\}$ indicates if the message is the preceding message of the target message.  Our goal is to learn a disentanglement model, which could predict which message in the context range is the preceding message of the target message by minimizing the loss from a given dataset $D$.

\section{Methodology}
 In this section, we propose our pre-trained DialBERT model to address the disentanglement problem.

\subsection{Domain adaptation}
We use the Ubuntu forum data to post-train BERT \cite{devlin2018bert}, in order to further improve our model performance (details shown in Appendix~\ref{adapt}). 
   
   To optimize the model, both Masked Language Model likelihood (MLM) and Next Sentence Prediction (NSP) are used as the post-train tasks, and the final loss function is formulated as
   \begin{equation} 
         Loss_{post-training} = Loss_{MLM} + Loss_{NSP}.
   \end{equation}

   \begin{equation} 
         L = -\frac{1}{N}(\sum\limits_{j\in N} y_{j}log{p_{j}} + \alpha(\sum_{i\in N}c_{i}log{p_{i}}))
   \end{equation}
   
    \begin{figure*}
      \includegraphics[width=17cm]{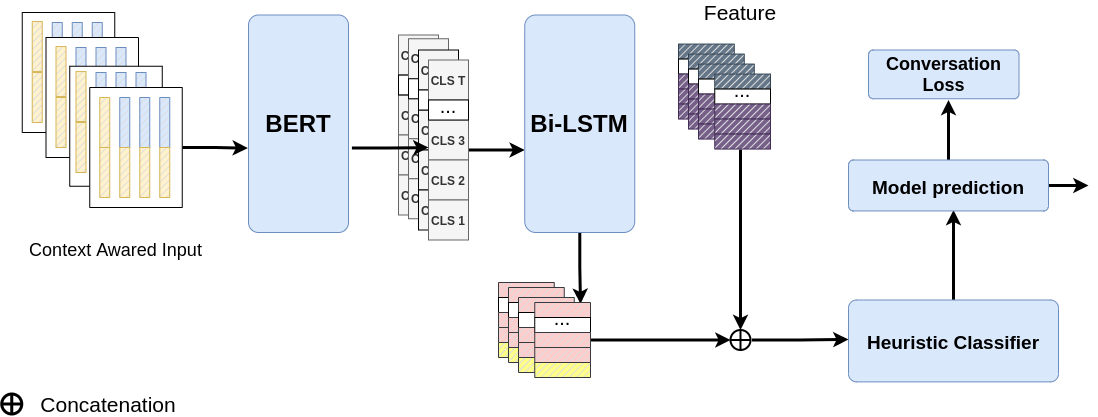}
      \caption{The overall architecture of our proposed model.}
      \label{fig2}
    \end{figure*}

  \subsection{DialBERT}

  We propose DialBERT to calculate the similarity between the target message and context messages.
  
\subsubsection{Context Aware Input}
  
To determine if two messages belong to the same conversation, context should be considered. For every message, we will take $T-1$ messages occurring before the target message in the same channel as its context, and combine the target message to each of its context as input. The similarity score is calculated between the target message and every context message. We concatenate them as follows:
   \begin{equation}
     Input_{i} = \left [cls,m_{1},m_{2}..,sep , c^{i}_{1}, c^{i}_{1},..,sep           \right ],
   \end{equation}
   where $i \in [0,1, ...,T-1]$ is the index of the context message. $cls$ and $sep$ are the start and separation tokens predefined in BERT. Note that $C_{0}$ is also the target message.

\subsubsection{Context BERT module}
BERT can encode a sentence with length less than 512 sub-words, which restricts the application of the model when adapted to downstream tasks. We propose a new model called Dialogue BERT(DialBERT) shown in Figure\ref{fig2}, which is capable of disentangling a conversation. Every time we package $T$ message pairs(target message with context messages) as one sample. After being encoded by BERT, we use a single layer Bi-LSTM to capture the information given by the encoded output, which means context information will be considered when DialBERT calculates the similarity of message pairs. In addition to BERT, RNN-based Bi-LSTM can help capture semantics across different message pairs. The output $f$ is formulated as:
   \begin{flalign}
     e_{i} &= \textbf{BERT}(Input_{i}), \\
     f_{i} &=  \textbf{BiLSTM}(e_{i}), \\
     \forall & i  \in [0,1, ...,T-1],
  \end{flalign}   
   where  $f_{i} \in \mathbb{R}^{H}$ is the similarity feature vector between the target message and context message with index $i$. Note that $f_{0}$ would be the similarity feature vector calculated between two target messages, which will be used as a ``pivot'' for classification.

  \subsubsection{Heuristic Classifier}
  In order to model higher-order interaction between the target message
  and context messages, we also compute the difference and element-wise product for the target similarity feature vector ($f_{0}$)
  and every corresponding context similarity feature vector ($f_{i}$). We duplicate target similarity feature vectors to form a feature matrix($F^{t} \in \mathbb{R}^{T\times H}$) having the shape of the context similarity feature matrix($F^{c}\in \mathbb{R}^{T\times H}$).
  Along this direction, we also further model this interaction by feeding tuple $G$ into feedforward neural networks with $Tanh$ activation function.
  \begin{align}
    F^{t}  &=[f_{0}, f_{0}, f_{0},..,f_{0}],\\
    F^{c}  &=[f_{0}, f_{1}, f_{2},..,f_{T-1}],\\
    G &= [F^{t}, F^{c}, F^{t} \cdot F^{c}, F^{t} - F^{c}], \\
    K &= \textbf{Tanh}(\textbf{G} \times \textbf{W}^\top + \textbf{b}), \\
    P &= \textbf{Softmax}(K). 
  \end{align}
$ P \in \mathbb{R}^{T}$ is the similarity score between target message and each context message. We adopt cross-entropy loss for DialBERT.

\subsection{Model}
    \begin{table*}
      \centering
      \setlength{\tabcolsep}{0.9mm}{
      \begin{tabular}{lcccccc}
      \hline
                                       &  VI     & ARI    & 1-1       & F1     & P      & R  \\ \hline
        Linear                         &  88.9   & -      & 69.5      & 21.8   & 19.3   & 24.9    \\
        Feedforward                    &  91.3   & -      & 75.6      & 36.2   & 34.6   & 38.0     \\
         $\times$ 10 union             &  86.2   & -      & 62.5      & 33.4   & 40.4   & 28.5     \\
         $\times$ 10 vote              &  \textbf{91.5}   & -  & \textbf{76.0}  & \textbf{38.0}     & 36.3   & \textbf{39.7}     \\
         $\times$ 10 intersect         &  69.3   & -      & 26.6      & 32.1   & \textbf{67.0}   & 21.1     \\
        Elsner(2008)                   &  82.1   & -      & 51.4      & 15.5   & 12.1   & 21.5     \\ 
        Lowe(2017)                     &  80.6   & -      & 53.7      & 8.9    & 10.8   & 7.6     \\ \hline
        
    Decom. Atten.~\cite{parikh-etal-2016-decomposable} (dev)* &  70.3   & -      & 39.8      & 0.6    & 0.9  & 0.7     \\
    Decom. Atten. + feature (dev)*                  & 87.4    & -      & 66.6      & 21.1    & 18.2  & 25.2     \\
    ESIM ~\cite{Chen2017EnhancedLF} (dev)*  &  72.1   & -      & 44.0      & 1.4   & 2.2   & 1.8     \\
    ESIM + feature (dev)*                  & 87.7    & -      & 65.8      & 22.6    & 18.9  & 28.3     \\
    BERT (dev)*                          & 74.7  & - & 45.4   & 2.2   & 2.6   & 2.7      \\
    BERT + feature (dev)*                      & 89.5  & -    & 71.7      & 21.4  & 30.0  & 25.0      \\
    MHT~\cite{DBLP:conf/aaai/ZhuNWNX20} (dev)*          &  82.1   & -      & 59.6      & 8.7    & 12.6   & 10.3  \\
    MHT +feature (dev)*           &  89.8   & -      & 75.4      & 35.8    & 32.7   & 34.2     \\
    DialBERT (dev)    &  \textbf{94.1}   & \textbf{81.1}      & \textbf{85.6}   & \textbf{48.0}    & \textbf{49.5}   &\textbf{46.6}     \\ \hline
        BERT                           &  90.8  & 62.9    & 75.0      & 32.5  & 29.3  & 36.6     \\
        BERT + feature                      &  92.2  & 65.9    & 76.8      & 37.8  & 33.9  & 42.5     \\
        DialBERT            &  92.6 & 69.6 & 78.5 & 44.1  &\textbf{42.3} &46.2  \\
        ConBERT + cov       &  \textbf{93.2} & \textbf{72.8} & \textbf{79.7} & \textbf{44.8}  &42.1 &\textbf{47.9}  \\
        DialBERT + feature             &  92.4  & 64.6      & 77.6      &42.2     & 38.8  & 46.3  \\
        DialBERT + future              &  92.3  & 66.3    & 79.1      & 42.6    &  40.0   & 45.6     \\
 \hline
      \end{tabular}}
      \caption{Results on Ubuntu test set, our work substantially outperforms prior work by all evaluation metrics. Six types of metrics are considered including the modified Variation of Information (VI) in (Kummerfeld et al. 2019), Adjusted rand index(ARI), One-to-One Overlap (1-1) of the cluster (Elsner and Charniak 2008), and the precision, recall, and F1 score between the cluster prediction and ground truth. Note that precision, recall, and F1 score are calculated using the number of perfectly matching conversations, excluding conversations with only one message (mostly system messages).}
      \label{tab5}
    \end{table*}
\subsubsection{Conversation Loss}

In the list of messages, different conversations are entangled together, and each conversation has its own semantic coherence and cohesion. 
Most previous studies failed to use the structure of each conversation when the parent message of a target message in the context is determined.
In order to encourage our model to find the parent message of the target message based on the context coherence of the conversation, we introduce conversation
loss in addition to the cross-entropy loss. In this way, our model can learn and leverage the structure of the conversation implicitly and will not suffer from a lack of conversation structure information during the inference/testing stage. 
The conversation loss is computed based on the matching score: 
  \begin{equation} 
        \mathcal{L}_{CV} = -\frac{1}{T}\sum_{i = 1}^{T} y_{i}^{c}\log({p_{i}}), 
  \end{equation}
  
    \begin{equation} 
            \mathcal{L}_{overall} =   \mathcal{L}_{CE} + \alpha  \mathcal{L}_{CV},
      \end{equation}
  where $\{y_{i}^{c}\}_{i=1}^{T}$ are the conversation labels and each $y_{i}^{c}$ is a binary label indicating whether the $i$-th context message is in the conversation same as the target message. $ \{ {p}_{i}\}_{i=1}^{T}$ are matching scores of message pairs. $T$ is context range.
   The intention of the conversation loss is to encourage the model to choose the parent message for a target message from the messages in the same conversation.

\section{Experimental Setup}
In this section, we will introduce the dataset we used to test our model and the setting of our model.
\subsection{Data}

In 2019, a conversation disentanglement dataset was published. This dataset contains 77563 messages of IRC manually annotated with reply-to relations between messages, which is proposed by \citep{kummerfeld-etal-2019-large}(details are shown in \ref{DATA}). Roughly 75K messages were used, which is still substantially higher than previous work.

In order to compare our work we employed several baseline models. Following the setting of \citep{kummerfeld-etal-2019-large}, we adopt linear and feedforward models as baselines. Aiming to import a strong baseline result, union, vote and intersect ensemble strategies given by \citep{kummerfeld-etal-2019-large} were used with feedforward model. To make our work more comparable,we also run the experiment using BERT (training details are shown in \ref{detail}). Moreover, we also including previous BRET based model's result proposed by \citep{DBLP:conf/aaai/ZhuNWNX20}. An extensive list of the modified pre-trained models are listed below.

\begin{itemize}
\item \textbf{BERT}  BERT model will be used to rank potential antecedents, and find preceding messages according to the ranking scores.
\item \textbf{BERT + feature} The same setting as BERT, combined with the same features used in Linear model\citep{kummerfeld-etal-2019-large}. Specifically, The features consist of three part: (1) Global-level features including year and frequency of the conversation. (2) Utterance level features including types of
message, targeted or not, time difference between the last message, etc., (3) utterance pair features including how far apart in position and time between the messages, whether one message targets another, etc.
\item \textbf{DialBERT} DialBERT with adaptation will be used to find preceding message according to the ranking scores.
\item \textbf{DialBERT} DialBERT with conversation loss.
\item \textbf{DialBERT + feature} Same linguistic features will be used as \textbf{BERT+feature}.
\item \textbf{DialBERT + future context} In this setting, Not only $T - 1$ messages occurred before this target message will be considered, $K$ messages occurred after target message will be considered as well during ranking potential antecedents.
\end{itemize}

\subsection{Training Details}
\label{detail}
We use the base version of BERT. The initial learning rate was set to be 2e-5. The maximum sequence length was set to 100.
The number of hidden unit $k=384$.
For the Conversation losses, the best hyperparameters are $\alpha = 0.1$.
Dropout is implemented in the output layer of the \emph{ConBERT} and Heuristic classifier with a ratio of $0.1$.
For the IRC dataset, we use a batch size of 4; the context range $T$ was set to 50. All code is written in the TensorFlow framework \cite{abadi2016tensorflow} and is trained on RTX TITAN (24G). We  publish our source code to help replicate our results \footnote{https://github.com/TeddLi/Disentangle}.

\subsection{Evaluation Metrics}
For the IRC dataset, we follow the setting in \citet{kummerfeld-etal-2019-large}. The evaluation metrics used in our experiments include: the modified Variation of Information (VI) \cite{kummerfeld-etal-2019-large}, Adjusted Rand Index (ARI), One-to-One Overlap (1-1) of the cluster~\cite{elsner-charniak-2008-talking}, as well as the precision, recall, and F1 score between the cluster prediction and ground truth. Note that the precision, recall, and F1 score are calculated using the number of perfectly matching conversations, excluding conversations that have only one message (mostly system messages). We take VI as the main metric.
For the Reddit dataset, we follow the setting of~\citet{DBLP:conf/aaai/ZhuNWNX20}. Specifically,  the
graph accuracy and the
conversation accuracy are adopted. The
graph accuracy is used to measure the average agreement between
the ground truth and predicted parent for each utterance.
The conversation accuracy is used to measure the average agreement between conversation structures and predicted structures. Specifically, only if all messages in a conversation are predicted correctly, the predicted structure is regarded as correct. We take graph accuracy as the main metric.

\section{Result \& Analysis}


\subsection{Overall Performance}    
Results are shown in Table~\ref{tab5}.
For previous work, an ensemble of $10$ feedforward models obtained through a vote were capable of reaching the best performance in 4 out of 5 evaluation metrics.

Different from other NLP tasks, BERT  does not perform well, which indicates semantic knowledge learned from pre-training is not a direct indicator of improvement for disentanglement.  
Post-combination with linguistic features, BERT does have advantages compared with the feedforward neural network according to evaluation metrics, which indicates semantic knowledge acquired by pre-training assists a model in performing better in downstream tasks through proper usage. Another reason might be that linguistic features import information across different sentences, which in turn assist BERT in making better decisions.

The result that DialBERT outperforms BERT on all 6 evaluation metrics could be explained by the vital importance of context in disentangle conversations, and DialBERT makes better use of pre-trained knowledge.
For DialBERT, external linguistic features only led to a slight improvement on a single metric, wheres the performance did not improve across other metrics. The reason might be DialBERT is powerful enough to capture most of the information given by those features.
Improvements are not shown by importing future context, which maybe attribute to the information acquired by precedent context being sufficient for DialBERT.
Dialogue BERT outperforms all previous models on all six evaluation metrics.

  \begin{table}
  \centering
  \setlength{\tabcolsep}{1.2mm}{
  \begin{tabular}{lccccc}
  \hline
                   &  VI  &  ARI & F1 & P & R  \\ \hline
   Model-AVG            &  \textbf{92.8}  & \textbf{72.1} & 43.3 & 39.8  & 47.6     \\
   Probability-AVG       &  92.6 & 66.6 & \textbf{45.3} & \textbf{42.1} & \textbf{49.0}      \\
   Vote-AVG                 &  92.6 & 66.6 & 45.0 & 42.0 & 48.5      \\ \hline
  \end{tabular}}
  \caption{Results for ensemble models on test set.}
  \label{tab4}
\end{table}  
In order to find out how well our model could handle the disentanglement problem, 
We also propose several DialBERT ensembles, the results is given on table ~\ref{tab4}. Three strategies were put into use, probability-average (Prob-AVG) , vote-average (Vote-AVG), model weight-average (Model-AVG) are put into use (details shows in \ref{Ensemble_strategy}).
The result shows that
Prob-AVG and Model-AVG strategies could reach better performance in comparison to vote-average. Model-AVG performs better in 1-Scaled VI and ARI metrics, Prob-AVG performs better in F1, recall and precision metrics.

Even though we propose DialBERT ensembles through post-training method, the best model could only reach an F1 score of 45.3\%, which indicates that the task of disentanglement is still a hard problem to solve.

\subsection{Ablation Results}   
Table\ref{tab6} displays the ablation analysis of different components.
As we discussed before, DialBERT has learned enough information to differentiate messages, so linguistic features do not result in a drastic improvement.
Unsurprisingly, the performance of the model drops in 5 out of 6 evaluation metrics after removal of post-training process, which indicates post-training learns useful semantic information, especially under the condition that the dataset is in a specific domain. After the removal of BiLSTM, results 
fall remarkably according to all evaluation metrics. In that way, the model has to make a prediction without any context consideration. As we discussed before, context is very important to disentangle a conversation. 
We can see from the ablation results, every modification to BERT in our model as mentioned in Table\ref{tab6} contributes to the final result, especially the BiLSTM component.

    \begin{table}
      \centering
      \setlength{\tabcolsep}{1.2mm}{
      \begin{tabular}{lcccccc}
      \hline
                                       &  VI     & ARI    & 1-1       & F1     & P      & R  \\ \hline
        Our model                  & \textbf{92.9}    & 68.1      & \textbf{80.0}     & 43.9   & 40.5  & \textbf{47.9}     \\
        - feature                  & 92.7    & \textbf{69.2}     & 78.4     & \textbf{44.3}    & \textbf{42.1}   & 46.7      \\
        - adaptation               & 92.5    & 67.8     & 78.6     & 41.0     & 37.6   & 45.1     \\
        - BiLSTM                    &  90.8  & 62.9    & 75.0      & 32.5  & 29.3  & 36.6\\
 \hline
      \end{tabular}}
      \caption{Ablation analysis of different components.}
      \label{tab6}
    \end{table}

\section{Conclusion}
Conversation disentanglement is a hard problem with broad application prospects.
In this paper, we propose a novel framework towards dealing with disentanglement. Different from previous work, we introduce both local and global semantics to disentangle conversations. Moreover, in order to make DialBERT perform well in the Ubuntu domain, we also post-train our model. Our model reaches state-of-the-art results on the newly published disentanglement dataset with a substantial margin in comparison to other baseline models.

\newpage

\bibliography{acl2020}
\bibliographystyle{acl_natbib}

\clearpage
\appendix
\section{Appendices}

\subsection{Related Work}

Pre-trained models such as BERT\cite{devlin2018bert} have been proven to work efficiently in many NLP tasks, but there is no previous work relevant to conversation disentanglement problem that uses pre-trained model. In this work we will present our own DialBERT model, which to best of knowledge is the first attempt to adopt a pre-trained model for disentanglement.

Domain adaptation(post-training) is an effective way to improve the performance of a model on a domain-specific corpus. Domain adaptation were first researched by \cite{Daume_III_2006}. Labeled data from both the source domain and target domain are needed for most adaptation work\cite{jiang2007instance,whang2019domain,howard2018universal,han2019unsupervised}. Recently, domain adaptation has also been used for multi-turn response selection in a retrieval-based dialog system by \cite{whang2019domain}. The dataset used in our work is derived from Ubuntu Internet Relay Chat(IRC). In order to make our model capture semantics related to Ubuntu, we used the external knowledge from DSTC8 \cite{kim2019eighth} to post-train our DialBERT.

Simultaneous conversation occurs not only in informal social interactions but also in conversations involving several participants in our daily lives. Aiming to separate intermingled messages into detached conversations, conversation disentanglement is the key research towards dealing with that problem.
The research for conversation disentanglement dates back to \cite{aoki2003mad} which conducted a study of voice conversations among 8-10 people with an average of 1.76 activate conversations at any given time. Further research not only propose datasets\cite{elsner-charniak-2008-talking,mehri2017chat,riou:hal-01698147} but also models \cite{mehri2017chat,jiang2018learning}. In recent studies, \citet{mehri2017chat} used recurrent neural networks(RNNs) to model adjacent messages as additional features. \citet{jiang2018learning} was the first to propose to use convolutional neural networks to estimate the conversation-level similarity between closely posted messages.
In recent research, \citet{DBLP:conf/aaai/ZhuNWNX20} proposed masked hierarchical transformer based on BERT to learn a better conversation structure. Their approach need conversation structure in advance to formulate the mask for their transformer model. During the test process, they construct their mask according to previous prediction, in that way, their mask could not give reliable conversation structure information, because of the relative low accuracy of the model's prediction.

    \begin{table}
      \centering
      \setlength{\tabcolsep}{1.5mm}{
      \begin{tabular}{lccc}
      \hline
                       & Train  &  dev & test  \\ \hline
      Speaker          &  15753  & 1220 & 1480    \\
      Conversation     &  17619  & 749 & 962  \\
      Messages         &  67463 &  2500 & 5000 \\
       \hline
      \end{tabular}}
      \caption{Statistic of disentanglement dataset that we used.}
      \label{tab2}
    \end{table}

Our model could introduce both global and local conversation semantics without explicitly introduce conversation structure result in achieving state-of-the-art results by outperforming other models by wide margin gap.

    \begin{table*}
      \small
      \centering
      \setlength{\tabcolsep}{1.0mm}{
      \begin{tabular}{lll}
      \hline
        Previous message index & Index  & Message  \\\hline
        ...                 & ...    & ... \\
        996                &1000    &  [03:04] Amaranth: @cliche American \\
        992                &1001    & [03:04] Xenguy: @Amaranth I thought you were -- welcome mortal ;-)\\
        1000                & 1002    & [03:04] cliche: @ Amaranth, hahahaha \\
        1003                & 1003    & === welshbyte  has joined \#ubuntu   \\
        997                & 1004   &   [03:04] e-sin: no i just want the normal screensavers     \\
        995                & 1005    &  [03:04] Amaranth: @benoy Do you have cygwinx installed and running?      \\
        1006                & 1006    &  [03:04] babelfishi: can anyone help me install my Netgear MA111 USB adapter? \\
        1004                & 1007     & [03:04] e-sin: i have a 16mb video card \\
        1008                & 1008     & === regeya  has joined \#ubuntu \\
        1007                & 1009     & [03:04] e-sin: TNT2 :) \\
        1001                & 1010     & [03:05] Amaranth: @Xenguy hehe, i do side development \\
        1007                & 1011     & [03:05] jobezone: @e-sin then it's xscreensaver and xscreensave-gl for opengl ones. \\
        1005                & 1012     & [03:05] benoy: how do i install that?  I couldn't find that in the list of things \\
        1010                & 1013     & [03:05] Amaranth: @Xenguy things like alacarte and easyubuntu \\
        ...                 & ...      & ...

      \end{tabular}}
      \caption{Dataset format.}
      \label{tab1}
    \end{table*}

\subsection{Domain Adaptation}
\label{adapt}
 \begin{figure}
  \includegraphics[width=7cm,height=3.5cm]{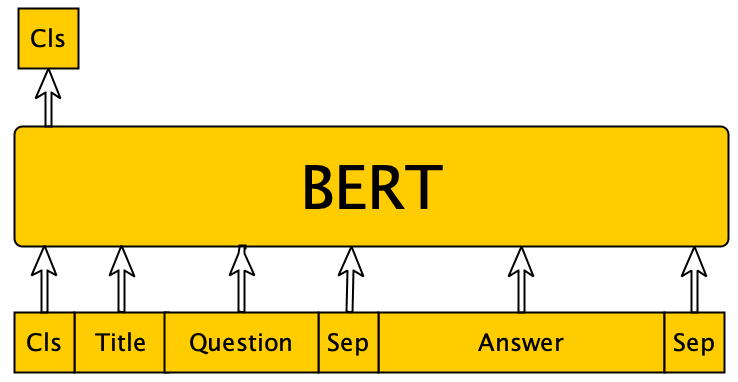}
  \caption{ Adaptation process}
  \label{fig1}
\end{figure}
Ubuntu forum data is given by Google DSTC8 \cite{kim2019eighth} competition track2's external source. Each sample contains a title and a question followed by an answer for that question. All questions lie within the context of Ubuntu. 
Different from random shuffling the external data sentences as input, we manually set the input of post-training process. As shown in figure\ref{fig1}, we followed the input format of BERT. Specifically, we used title and question as sentence A, and answer as sentence B. For every sentence A, we randomly pick a answer as a negative sample.

\subsection{Dataset}
\label{DATA}
The dataset proposed by \citep{kummerfeld-etal-2019-large} is 16 times larger than all previously released datasets combined. Here is the data sample in table \ref{tab1}. 
The training set was sampled in three ways: (1) 95 uniform length samples, (2) 10 smaller samples to check annotator agreement, (3) 48 time spans of one hour, which could maintain the diversity in terms of number of messages, the number of participants, and the percentage of directed messages. 
Moreover, with the use of \textbf{Adjudication}, the quality of this dataset is reliable.
Practically, we will use train, dev, test folders of that dataset, excluding leave Pilot, Channel folders out. 
 As shown in table \ref{tab2}, about 74963 messages were used, which is still much larger than previous work.

  \begin{figure}
      \includegraphics[width=8.5cm]{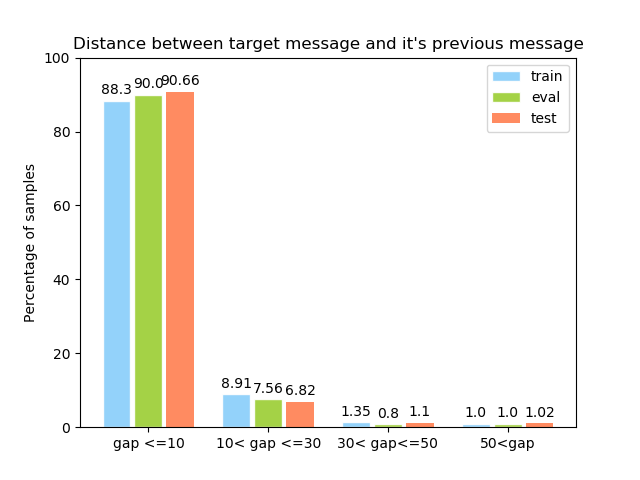}
      \caption{The percentage of Distance between target message and its preceding message}
      \label{fig3}
    \end{figure}

\subsection{Training Detail}
\label{detail}
We use the base version of BERT. The initial learning rate was set to 2e-5. The maximum sequence length was set to 100. We use a batch size of 4. As shown in figure \ref{fig3}, around 99\% target messages could find their respective preceding messages in 50 round context, so context range $T$ was set to 50, with 49 messages before target message were considered as context. DialBERT + future context will consider 10 messages occurring after the target message. The hidden size of the Bi-LSTM component of our model was set to 384, so the concatenated output is 768 which is the same as \textit{base-BERT}'s output. The heuristic classifier has 3072 hidden units. 
Dropout is implemented to the output of DialBERT and Heuristic classifier with the ratio of $0.1$.
\subsection{Ensemble Strategy}
\label{Ensemble_strategy}
    \begin{itemize}
  \item{\textbf{Model-AVG}}
  In Model-AVG process, we average the weight of the model across several Dialogue BERT models.
  \item{\textbf{Prob-AVG}}
  In Probability-AVG process, we average the weight of the model prediction probability for each sample across different model.
  \item{\textbf{Vote-AVG}}
  In Vote-AVG process, we create an ensemble of DialBERT by considering the context message with most votes from each model as our prediction within the same conversation with the target message.
    \end{itemize}
    Specifically, different strategy need different amount of models to reach best result. For Model-AVG is 2 models. Probability-AVG is 8 models. Vote-AVG is 8 models.
\end{document}